\documentclass{article}

\usepackage[preprint]{neurips_2026}

\usepackage[utf8]{inputenc} % allow utf-8 input
\usepackage[T1]{fontenc}    % use 8-bit T1 fonts
\usepackage{hyperref}       % hyperlinks
\usepackage{url}            % simple URL typesetting
\usepackage{booktabs}       % professional-quality tables
\usepackage{amsfonts}       % blackboard math symbols
\usepackage{nicefrac}       % compact symbols for 1/2, etc.
\usepackage{microtype}      % microtypography
\usepackage{xcolor}         % colors
\usepackage{graphicx}
\usepackage{subcaption}  % For subfigure environment
\usepackage{amssymb}
\usepackage{amsthm}
\usepackage{amsmath}
\usepackage{todonotes}

\title{Beyond Red-Teaming: Formal Guarantees of LLM Guardrail Classifiers}

\author{%
  Nikita Kezins \\
  Delft University of Technology \\
  \texttt{nikitakezins@gmail.com} \\
  \And
  Urbas Ekka \\
  \texttt{urbasekka@gmail.com} \\
  \And
  Pascal Berrang \\
  University of Birmingham \& Zeroth Research \\
  \texttt{p.p.berrang@bham.ac.uk} \\
  \And
  Luca Arnaboldi \\
  University of Birmingham \& Zeroth Research\\
  \texttt{l.arnaboldi@bham.ac.uk} \\
}

\begin{document}

\maketitle

\begin{abstract}
Guardrail Classifiers defend production language models against harmful behavior, however, although results may seem promising in testing, they provide no formal guarantees. 
Unfortunately, providing formal guarantees for such models is hard because ``harmful behavior'' has no natural specification in a discrete input space: and the standard $\epsilon$-ball properties used in other domains do not carry semantic meaning. 
We close this gap by shifting verification from the discrete input space to the classifier's pre-activation space, where we define a harmful region as a convex shape enclosing the representations of known harmful prompts.
Because the sigmoid classification head is monotonic, certifying the worst-case point of this region is sufficient to certify the entire region, yielding a closed-form soundness proof without approximation in $O(d)$ time.
To formally evaluate these types of classifiers, we propose two constructions of such regions: SVD-aligned hyper-rectangles, which yield exact SAT/UNSAT certificates, and Gaussian Mixture Models, which yield probabilistic certificates over semantically coherent clusters. 
Applying this framework to three author-trained Guardrail Classifiers (BERT, GPT-2 and Llama-3.1-8B) on the toxicity domain, every hyper-rectangle configuration returns SAT, exposing verifiable safety holes across all classifiers, despite seemingly high empirical metrics (F1, recall, etc). 
Probabilistic GMM certificates also expose a divergent structural stability in how these models represent harm. While GPT-2 and Llama-3.1-8B maintain robust coverage of 90\% and 80\% across varying boundaries, BERT's safety guarantees prove uniquely volatile. This `coverage collapse' to 55\% at the optimal threshold ($\tau^*$) reveals a sparsely populated safety margin in BERT, which only achieves full coverage by adopting an extremely conservative pessimistic threshold.
These approaches combined, provide new insights on how effective Guardrail Classifiers really are, beyond traditional red-teaming.
\end{abstract}

\section{Introduction}
\label{sec:introduction}

The rapid integration of Large Language Models (LLMs) into production environments has intensified the need for robust safety guardrails. Among the most prominent defenses are Guardrail Classifiers -- an umbrella term encompassing rule-based systems~\cite{rebedea2023nemoguardrailstoolkitcontrollable}, input-output safeguards~\cite{inan2023llamaguardllmbasedinputoutput}, and state-of-the-art Constitutional Classifiers~\cite{sharma2025constitutionalclassifiersdefendinguniversal, cunningham2026constitutionalclassifiersefficientproductiongrade}. These models serve as specialized safety filters to detect and intercept adversarial jailbreaks.
These classifiers demonstrate strong empirical resilience against complex multi-turn attacks and universal adversarial triggers.
However, their safety claims rest exclusively on red-teaming results, and red-teaming establishes only the absence of attacks tried, not the absence of attacks possible.

Formal verification offers a complementary path by providing mathematical guarantees over entire input regions.
Applying it to Transformer-based classifiers, however, runs into two well-known obstacles.
First, standard verification techniques such as Satisfiability Modulo Theories (SMT) or linear relaxation-based methods \cite{katz2017reluplexefficientsmtsolver} methods do not scale to the depth and non-linearity of modern Transformers; as depth grows, approximation errors accumulate and the resulting bounds become too loose to be informative~\cite{gowal2019,li2023sokcertifiedrobustnessdeep}.

Second, the discrete nature of language offers no analogue of the $L_\infty$-balls used in computer vision: a tractable, semantically meaningful neighbourhood in token space remains an open problem.
Together, these obstacles have left a clear gap -- no prior work has formally verified whether a Guardrail Classifier actually enforces its constitution.

We close this gap with the first formal safety analysis of Guardrail Classifiers, replacing attack-success rates with mathematically provable certificates.
Our approach sidesteps both obstacles by relocating verification.
Rather than propagating bounds through the full Transformer, we verify only the classification head on the last hidden state.
Rather than defining neighbourhoods in token space, we define safety regions directly in this representation space, where semantically similar inputs cluster geometrically.
This approach leverages a key architectural insight: the monotonicity of the sigmoid activation function, which permits exact, closed-form analysis of whether a given safety region lies entirely above the classifier's threshold -- UNSAT, the region is classified correctly -- or admits the existence of a counterexample -- SAT, the classifier has a safety hole within the specification.

This formulation lets us pose three questions about Guardrail Classifiers in a falsifiable way. \textbf{RQ1:} Do Guardrail Classifiers trained on realistic safety data exhibit verifiable safety holes over semantic regions of harmful behavior? \textbf{RQ2:} How does specification fidelity and certified coverage vary across classifiers and model scales? \textbf{RQ3:} What are the trade-offs between deterministic (hyper-rectangles) and probabilistic (GMM) specifications, and how does threshold selection affect the strength of the resulting certificate?

Our contributions are three-fold:

\textbf{A closed-form gap-analysis framework for Guardrail Classifiers:} We prove that, for classifiers with a sigmoid head, deciding whether a region bounded by a hyper-rectangle lies entirely above the classification threshold reduces to evaluating the head at a single worst-case point. The complexity of the procedure is linear for the verification of each hyper-rectangle ($O(d)$).

\textbf{Two complementary specifications of harm:} We adapt the hyper-rectangle approach introduced in ANTONIO~\cite{casadio2023antoniosystematicmethodgenerating} for deterministic SAT/UNSAT certificates and introduce a Gaussian Mixture Model construction for probabilistic certificates over semantically coherent clusters of harm.

\textbf{Empirical safety gap analysis across classifiers:} Applying our framework to three author-trained Guardrail Classifiers (BERT, GPT-2 and Llama-3.1-8B) on the toxicity domain, every hyper-rectangle configuration returns SAT, confirming safety holes across all classifiers. GMM certificates reveal substantial variation in certified coverage -- GPT-2 (90\%), Llama-3.1-8B (80\%) and BERT (55\%) under the optimal threshold $\tau^*$ -- with BERT reaching 100\% under the pessimistic threshold $\tau_{pess}$.\footnote{Source code is available at \url{https://github.com/entfane/formal_verification_guardrails}.}

% \begin{itemize}
%     \item \textbf{A closed-form gap-analysis framework for Guardrail Classifiers:} We prove that, for classifiers with a sigmoid head, deciding whether a region bounded by a hyper-rectangle lies entirely above the classification threshold reduces to evaluating the head at a single worst-case point. The complexity of the procedure is linear for the verification of each hyper-rectangle ($O(d)$).
%     \item \textbf{Two complementary specifications of harm:} We adapt the hyper-rectangle approach introduced in ANTONIO~\cite{casadio2023antoniosystematicmethodgenerating} for deterministic SAT/UNSAT certificates and introduce a Gaussian Mixture Model construction for probabilistic certificates over semantically coherent clusters of harm.
%     \item \textbf{Empirical safety gap analysis across classifiers:} Applying our framework to three author-trained Guardrail Classifiers (BERT, GPT-2 and Llama-3.1-8B) on the toxicity domain, every hyper-rectangle configuration returns SAT, confirming safety holes across all classifiers. GMM certificates reveal substantial variation in certified coverage -- GPT-2 (90\%), Llama-3.1-8B (80\%) and BERT (55\%) under the optimal threshold $\tau^*$ -- with BERT reaching 100\% under the pessimistic threshold $\tau_{pess}$.\footnote{Code to reproduce all verification results is provided in the supplementary ZIP. An anonymized repository for review is available at \url{https://anonymous.4open.science/r/formal_verification_guardrails-0DED/README.md}}
% \end{itemize}

\section{Formal Preliminaries}
\label{sec:overview}

We now formalize the mathematical conditions required to certify a Guardrail Classifier, establishing the theoretical foundation for the two verification methods developed in Section~\ref{sec:method}. We consider a fixed Transformer backbone $\Phi: \mathcal{X} \to \mathbb{R}^d$ and a classification head $f: \mathbb{R}^d \to (0,1)$, where $f(\mathbf{x}) = \sigma(W^\top \mathbf{x} + b)$, $W \in \mathbb{R}^d$ are the head weights, $b \in \mathbb{R}$ is the bias, and $\sigma$ is the sigmoid activation. The core challenge is to verify that for a semantic region $\mathcal{R} \subset \mathbb{R}^d$ in the pre-classification activation space, the property $f(\mathbf{x}) > \tau$ holds $\forall \mathbf{x} \in \mathcal{R}$.

\noindent\textbf{Exact Bounds via Sigmoid Monotonicity.} \label{sec:verif_exact_bounds}
The primary advantage of verifying the classification head in isolation is the exploitation of its monotonic properties. Since $\sigma$ is strictly increasing, minimizing $f$ over $\mathcal{R}$ reduces to minimizing the pre-activation $z = W^\top \mathbf{x} + b$ over $\mathcal{R}$. In the case $\mathcal{R}$ is a hyper-rectangle $\mathcal{H}$ defined by lower bounds $\mathbf{l}$ and upper bounds $\mathbf{u}$, the minimum pre-activation value $z_{\min}$ is given by:

\begin{equation}
    z_{\min} = \sum_{i=1}^{d} \min(w_i l_i, w_i u_i) + b
\end{equation}

Because $\sigma(z)$ is strictly increasing, the global minimum of $f$ over $\mathcal{H}$ is $\sigma(z_{\min})$. A region $\mathcal{R}$ is certified as \textit{formally safe} if $\sigma(z_{\min}) > \tau$. This closed-form solution bypasses iterative solvers or SMT-based search entirely. The formal proof is provided in Appendix~\ref{app:sigmoid_proof}.

\noindent\textbf{Probabilistic Bounds via Gaussian Transformations.}
For probabilistic verification, because the classification head applies a linear transformation before the sigmoid, each component $k$ induces a univariate Gaussian over the pre-activation $z$:

\begin{equation}
    z \sim \mathcal{N}(W^\top \mu_c + b,\; W^\top \Sigma_c W)
\end{equation}

Rather than propagating distributions through the non-linear sigmoid, we map the threshold $\tau$ back into pre-activation space via the inverse sigmoid $\sigma^{-1}(\tau) = \ln\!\left(\frac{\tau}{1-\tau}\right)$ and compute the probability mass exceeding it:
\begin{equation}
    P(z > \sigma^{-1}(\tau)) = 1 - \Phi\!\left(\frac{\sigma^{-1}(\tau) 
    - \mu_z}{\sigma_z}\right)
\end{equation}
where $\Phi$ is the standard normal CDF. This yields a probabilistic safety certificate: the probability that a harmful activation drawn from the modelled distribution is correctly classified above the threshold.

\section{Method}
\label{sec:method}

We verify harmful regions defined within the last hidden state of the Transformer, immediately preceding the classification head. The final layer of a Transformer is heavily context-enriched, encapsulating the maximum information density within the aggregation token. By defining our specifications in this space, we capture data points that exhibit semantically meaningful relationships. Specifically, we propagate $N$ harmful samples through the Guardrail Classifiers, extracting the representations of the token that aggregates the entire prompt's features (e.g., the final token in decoder-only models or the initial token in encoder models). These $N$ samples yield a set of $d$-dimensional points, where $d$ denotes the hidden state dimensionality, representing the model's internal encoding of harmfulness. We utilize these points to formally define our harmful behavior specifications. Crucially, these specific samples serve as the empirical definition of harmful behavior, which in the context of our study is explicitly instantiated as toxicity.

\begin{figure}
    \centering
    \includegraphics[width=1\linewidth]{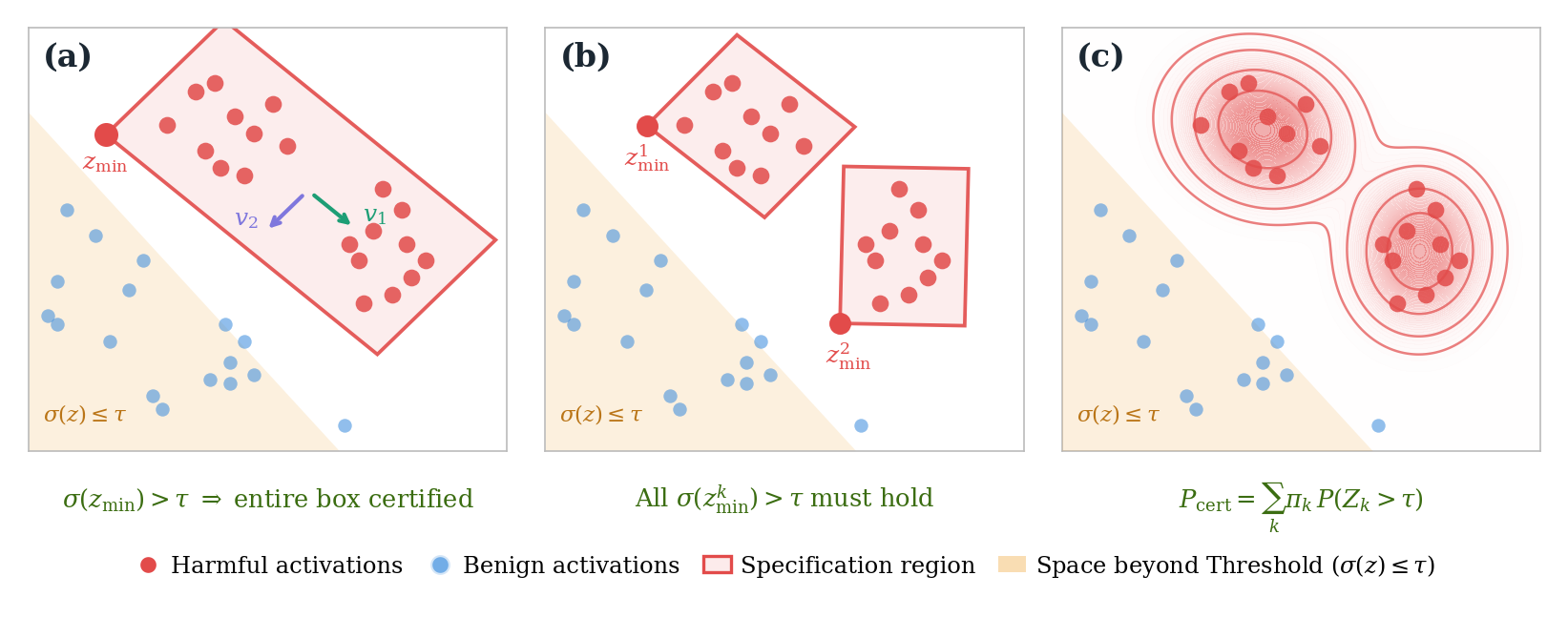}
    \caption{Illustration of Specifications: (a) Displays the Single Hyper-Rectangle construction with SVD rotation, (b) Displays Multiple Hyper-Rectangle construction clustered with HDBSCAN, (c) Displays GMM construction. The shaded region is where the model classifies activations as benign.}
    \label{fig:spec_illustr}
\end{figure}

\noindent\textbf{Construction of Semantic Regions.}
We define a harmful region as a bounded subspace of the pre-classification activation space constructed from the representations of known harmful inputs. The efficacy of our verification relies on the quality of this region $\mathcal{R}$. If $\mathcal{R}$ is too large, it may encompass safe activations; if too small, it lacks coverage. We formalize our two construction methods below (illustrated in Figure~\ref{fig:spec_illustr}).

\noindent\textbf{Single Hyper-rectangle Specification.}
Given the $N$ points extracted from the pre-classification layer, we construct a bounding hyper-rectangle. To minimize the empty-space problem in high-dimensional verification, we do not define $\mathcal{H}$ along the standard coordinate axes. For each cluster of harmful activations, we use Singular Value Decomposition (SVD) to obtain a rotation matrix $V^\top$ aligned with the principal axes of the data, allowing a tightly fitted bounding box to be drawn around the cluster. Crucially, the same rotation matrix $V^\top$ is applied to the classifier weights $W$, yielding rotated weights $\tilde{W} = V^\top W$. Because the classification score is computed as an inner product $W^\top x$, and rotation is an orthogonal transformation ($V^\top V = I$), we have $W^\top x = \tilde{W}^\top \tilde{x}$ for any rotated point $\tilde{x} = V^\top x$. This means verification in the rotated space is not an approximation -- it is algebraically identical to verification in the original space. The SVD rotation therefore serves a purely geometric purpose: it minimizes the volume of the bounding hyper-rectangle, tightening the specification without introducing any error into the verification result. For a visual representation, see Figure~\ref{fig:spec_illustr}(a).
By identifying the minimum and maximum values along each dimension we finalize the geometric bounds. Verification of the resulting region follows directly from the closed-form procedure established in Section~\ref{sec:verif_exact_bounds}: we evaluate $\sigma(z_{\min})$ and check whether it exceeds the classification threshold $\tau$. The complete formal proof is provided in Appendix~\ref{app:sigmoid_proof}.

% Given the $k$ points extracted from the pre-classification layer, we construct a bounding hyper-rectangle by identifying the minimum and maximum values along each dimension in the SVD-rotated space, following the construction described in Section~\ref{sec:svd_aligned_hyperrect}. Verification of the resulting region follows directly from the closed-form procedure established in Section~\ref{sec:verif_exact_bounds}: we evaluate $\sigma(z_{\min})$ and check whether it exceeds the classification threshold $\tau$. The complete formal proof is provided in Appendix~\ref{app:sigmoid_proof}.

\noindent\textbf{Multiple Hyper-Rectangle Specification.}
Building upon the single bounding region formulation, we extend our approach to encapsulate multiple localized regions of the activation space. We achieve this by clustering the $N$ points using Hierarchical Density-Based Spatial Clustering of Applications with Noise (HDBSCAN~\cite{10.1007/978-3-642-37456-2_14}) equipped with a cosine distance metric to evaluate neighborhood densities. We select HDBSCAN over k-means primarily because the true number of localized harmful regions is unknown, and we require a method that does not force us to pre-define the number of clusters a priori. By leveraging a Minimum Spanning Tree (MST) and core distances that account for the relative density between points, this approach dynamically extracts clusters based on their local topology. Furthermore, we utilize cosine distance because it effectively captures semantic similarity independent of vector magnitude in high-dimensional spaces. By parameterizing the algorithm solely with a minimum cluster size, we can dynamically partition the data into multiple distinct clusters. Specifically, we equate the min\_samples parameter to this minimum cluster size, effectively defining the $m$ used to calculate the core distance. The minimum cluster size is treated as a sensitivity parameter; we report results across a range of values in Section~\ref{sec:mult_hyperrect_results} and select the configuration that maximizes specification fidelity while still producing more than one cluster. This density-based approach allows us to accurately capture multiple, highly concentrated regions within the activation space where representations of harmful behavior are localized. For each identified cluster, we construct a dedicated hyper-rectangle specification. The verification process naturally extends to this multi-region paradigm. 

\noindent\textbf{Gaussian Mixture Model.}
In contrast to deterministic hyper-rectangle bounds, we also investigate probabilistic verification methods by modelling our specification as a Gaussian Mixture Model (GMM). We model the harmful pre-classification activation space as a mixture of $K$ Gaussians:
\begin{equation}
    p(\mathbf{x}) = \sum_{c=1}^K \pi_c \mathcal{N}(\mathbf{x}; \mu_c, \Sigma_c)
\end{equation}
We fit $K$ Gaussian components to the $N$ points in the activation space, see Figure~\ref{fig:spec_illustr}(c). This approach allows us to model the covariance structure of the data, capturing complex, interconnected features among points assigned to the same Gaussian component. We apply a linear transformation to these Gaussian distributions. Because the subsequent sigmoid activation is non-linear, we operate in the pre-sigmoid space by applying the inverse sigmoid function to the decision threshold (calculations are detailed in the Appendix~\ref{app:prob_verification}).

\section{Evaluation}
\label{sec:evaluation}

In this section we evaluate our representational specifications and perform a formal safety gap analysis of Guardrail Classifiers. Our evaluation objective is to determine if current classifiers can provide formal guarantees over the semantic manifold of harm, or if they leave "safety holes" where harmful prompts could bypass the guardrail.

\subsection{Experimental Setup} \label{sec:exp_setup}
To evaluate and verify our specifications, we utilize a subset of the Toxigen~\cite{hartvigsen2022toxigenlargescalemachinegenerateddataset} dataset comprising 10,100 total exchanges (10,000 harmful and 100 benign). For our empirical evaluation of specification fidelity, we construct the boundaries using 9,900 harmful records and test them on a holdout set of the remaining 100 harmful and 100 benign records. However, for the final formal verification of the classifiers, we construct the safety regions using the entirety of the 10,000 harmful records to maximize semantic coverage. We measure the empirical fidelity of our geometric boundaries using the following criteria:

\textbf{Deterministic Shapes (Single and Multiple Hyper-rectangles):} We measure specification Recall (the frequency of harmful activations successfully captured within the bounded regions) and Precision (the frequency of benign activations correctly excluded from these regions).

\textbf{Probabilistic Shapes (Gaussian Mixture Models):} An activation is classified as "within the specification" if it resides above the 5th percentile of the GMM's probability density. This threshold defines the high-probability manifold of harm within the latent space.

% \begin{itemize}
% \item \textbf{Deterministic Shapes (Single and Multiple Hyper-rectangles):} We measure specification Recall (the frequency of harmful activations successfully captured within the bounded regions) and Precision (the frequency of benign activations correctly excluded from these regions).

% \item \textbf{Probabilistic Shapes (Gaussian Mixture Models)}: An activation is classified as "within the specification" if it resides above the 5th percentile of the GMM's probability density. This threshold defines the high-probability manifold of harm within the latent space.
% \end{itemize}

By evaluating the specs on unseen data, we assess their generalization capability, confirming that the verified regions capture subspaces strictly aligned with the constitutional definition of harm.

Furthermore, to test the robustness of these specifications, we conduct a sensitivity analysis. For the deterministic multi-region approach, we vary the minimum cluster size $m$ used in HDBSCAN. For the probabilistic approach, we vary the number of GMM components $K$. This allows us to identify the optimal trade-off between specification tightness and representational accuracy before applying them to formally verify our classifiers.

We also evaluate every specification under two thresholds, optimal and pessimistic. The optimal threshold $\tau^*$ is selected via Youden's J statistic, which provides the point that best balances true-positive rate against false-positive rate, which is the optimal decision boundary for the classifier. The pessimistic threshold $\tau_{pess}$ is set to the minimum classifier score on the evaluation set which still guarantees perfect recall. As the formal verification procedure yields exact closed-form certificates and GMM fitting is performed once on the full construction set, all reported results are deterministic and no error bars are applicable. 
Full details regarding the computational resources required to reproduce these experiments, as well as the licensing terms for all utilized datasets and base models, are provided in Appendix~\ref{app:reprod_asset_details}.

\subsection{Specification Fidelity Results}

\begin{figure}[t]
     \centering
     % First Subfigure
     \begin{subfigure}[b]{0.48\textwidth}
         \centering
         \includegraphics[width=\textwidth]{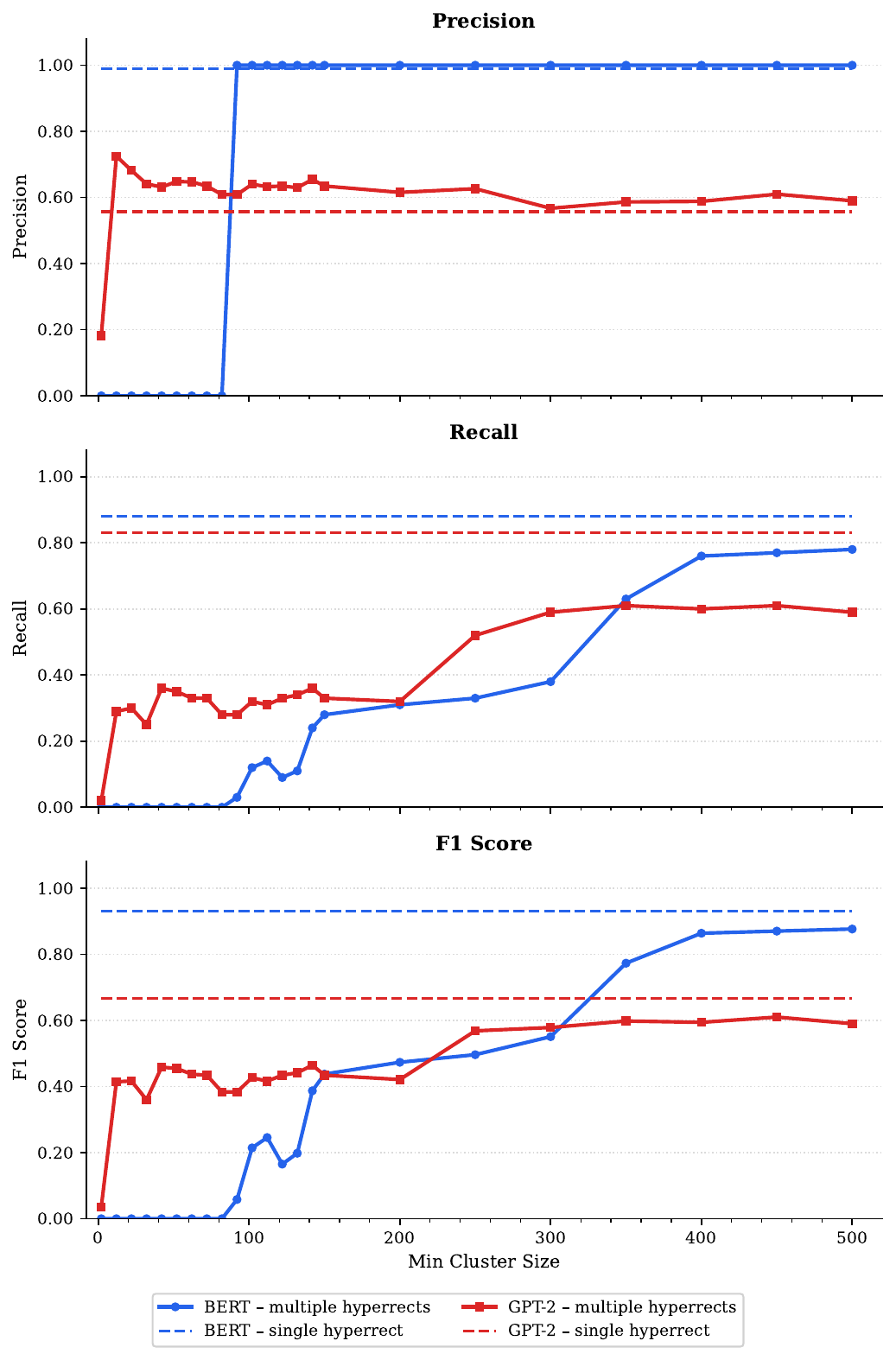}
         \caption{GPT-2 and BERT} % Adds the (a) marker
         \label{fig:mult_hyperrect_GPT2_BERT}
     \end{subfigure}
     \hfill
     % Second Subfigure
     \begin{subfigure}[b]{0.48\textwidth}
         \centering
         \includegraphics[width=\textwidth]{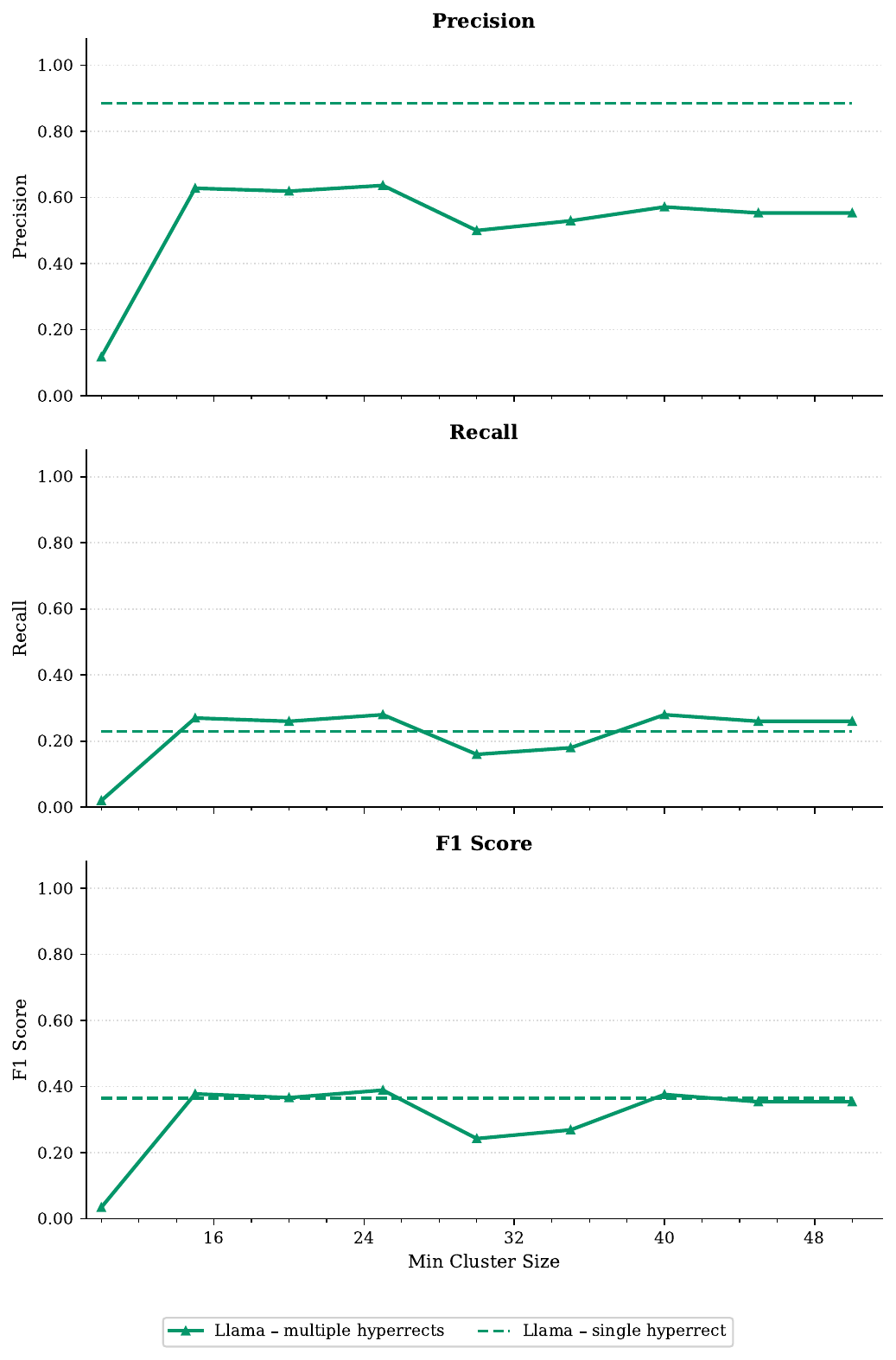}
         \caption{Llama-3.1-8B} % Adds the (b) marker
         \label{fig:mult_hyperrect_llama}
     \end{subfigure}
     \caption{Single and Multiple Hyper-Rectangle specification performance compared across Precision, Recall, and F1. The results highlight the relative clustering performance of smaller models (BERT, GPT-2) compared to Llama-3.1-8B.}
     \label{fig:mult_hyperrect}
\end{figure}

Before formally verifying the classifiers, we must assess the optimal trade-off between specification tightness and representational accuracy for our three geometric approaches.

\noindent\textbf{Single Hyper-rectangles.} \label{sec:single_hyprrect_results}
As detailed in Figure~\ref{fig:mult_hyperrect}, while a single SVD-aligned bounding hyper-rectangle provides the most straightforward path to formal verification, it inherently suffers from over-approximation. Smaller models (BERT and GPT-2) maintain relatively balanced F1-scores under this construction, indicating their harmful representations occupy a compact subspace. Conversely, Llama-3.1-8B exhibits a severe drop in specification recall to 23\%, while retaining a high specification precision of 89\%. This asymmetry reflects a structural property of the larger model: the classifier distributes its representations of harmful behavior across a vast and highly dispersed manifold. Consequently, while a tightly fitted geometric enclosure successfully isolates a localized cluster of harm (yielding high precision), it fundamentally fails to capture the vast expanse of the true harmful area, leading to the severe drop in recall.

\noindent\textbf{Multiple Hyper-rectangles.} \label{sec:mult_hyperrect_results}
To address the over-approximation, we use HDBSCAN with cosine similarity to partition harmful representations into localized sub-clusters, wrapping each in a tightly-fitted hyper-rectangle. As illustrated in Figure~\ref{fig:mult_hyperrect}, this introduces a scale-dependent precision-recall trade-off. For smaller models, finer partitioning yields modest precision gains. GPT-2 specification precision improves from 56\% to approximately 65\%, but specification recall falls from 83\% to roughly 61\%. BERT maintains specification precision near 100\%, but recall consistently falls short of the single hyper-rectangle baseline. Conversely, for larger models, the trade-off reverses. The multi-region approach yields a marginal specification recall gain, peaking at 28\%, while specification precision deteriorates from 89\% to approximately 64\%. This indicates Llama-3.1-8B's harmful activations are so geometrically dispersed that even local sub-clusters remain internally impure.

\begin{figure}[htbp]
     \centering
     \includegraphics[width=0.9\linewidth]{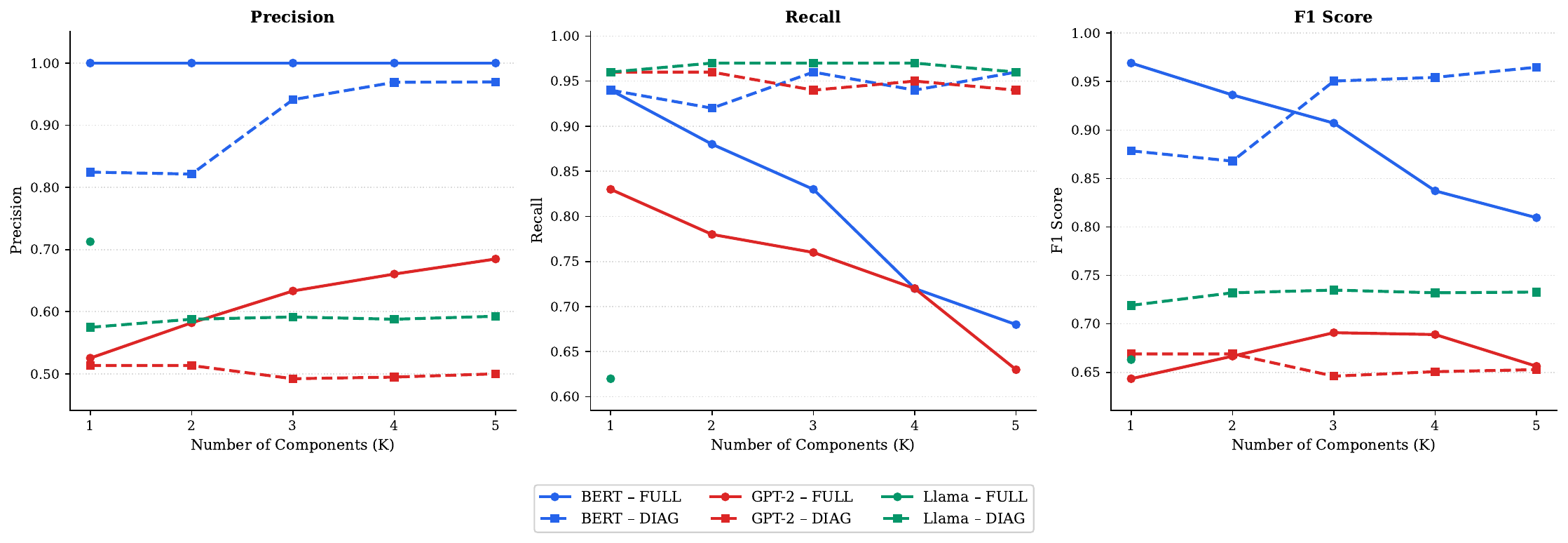}
     \caption{Performance of Gaussian Mixture Models (GMM) at the 5th percentile boundary. Precision, Recall, and F1-scores are evaluated across a varying number of components ($K$) for BERT, GPT-2 and Llama-3.1-8B. Llama-3.1-8B is restricted to a single data point for full covariance estimation, as the dimensionality of the hidden state ($d$) renders a multi-component evaluation computationally infeasible.}
     \label{fig:gmm_spec}
\end{figure}

\noindent\textbf{Probabilistic Boundaries.}
As Figure~\ref{fig:gmm_spec} demonstrates, GMMs provide the highest flexibility, revealing a distinct trade-off: full covariance yields higher precision, whereas diagonal covariance offers superior recall. Full-covariance GMMs achieve near-perfect precision for BERT across all evaluated components, though recall degrades as the number of components increases. Due to its high dimensionality, Llama-3.1-8B is restricted to a single component for full covariance, yielding modest baseline results. By shifting Llama-3.1-8B to a diagonal covariance structure, we observe a dramatic reversal: specification recall surges, peaking at an impressive 97\%. This underscores the flexibility of GMMs to prioritize either strict geometric tightness or broad semantic coverage simply by tailoring the covariance structure to the model's scale.

\noindent\textbf{Formal Verification of Guardrail Classifiers.}
Having established the fidelity of our specifications, we proceed to the formal verification of the Guardrail Classifiers. Leveraging our proof regarding the monotonicity of the sigmoid classification head, we evaluate whether any input activation residing within the defined safety specifications can bypass the classification threshold.
As shown in Table~\ref{tab:model_results}, our exact verification framework yields SAT results for both single and multiple hyper-rectangle configurations across all three architectures, regardless of whether the optimal ($\tau^*$) or pessimistic ($\tau_{pess}$) threshold is applied. SAT indicates that a counterexample exists, meaning we successfully identified "safety holes" within these tightly bounded harmful regions, where the classifier's output incorrectly falls below the toxicity threshold. This mathematically proves that none of the tested classifiers provide strict, deterministic safety guarantees over the semantic manifold of harm. 

Crucially, our verification operates exclusively on the pre-activation space of the final aggregation token. Because this single continuous vector represents a highly compressed encoding of the entire preceding context, it is fundamentally impossible to unspool a specific SAT coordinate back into a full, multi-token discrete sequence. Consequently, we cannot provide the exact textual prompts that correspond to the verified SAT counterexamples. However, to understand what caused the SAT results, we conducted an independent empirical investigation into the raw samples used to construct these regions. By running inference on the construction points, we found that several of these original, known-harmful prompts already yielded classifier scores significantly lower than the safety thresholds. This confirms that our SAT certificates are not mere artifacts of geometric over-approximation, but mathematically formalize real adversarial misclassifications already present within the bounds. Appendix~\ref{app:adv_examples} provides concrete examples: across the construction set, known-harmful prompts produce scores as low as 0.002--0.070 for all three classifiers, falling drastically below both $\tau^*$ and $\tau_{pess}$ (Table~\ref{tab:min_scores}). Representative exchanges are included to illustrate the nature of these empirical misclassifications. 

In contrast to the deterministic bounds, our probabilistic GMM certificates reveal a stark variance in safety guarantees that is highly sensitive to threshold selection. Under the optimal threshold ($\tau^*$), the classifiers successfully bound a large majority of the harmful distribution mass for GPT-2 (90\%) and Llama-3.1-8B (80\%), whereas BERT exposes a massive probabilistic safety gap, bounding only 55\% of the modelled mass. However, when evaluating under the pessimistic threshold ($\tau_{pess}$), certified coverage improves dramatically. BERT attains perfect 100\% probabilistic coverage, and GPT-2 rises to 98\%. Llama-3.1-8B's coverage remains stable at 80\%, as its optimal and pessimistic thresholds are identical.

\begin{table}[t]
  \centering
  \caption{Verification results ($\tau^* = \tau_{pess}$ for Llama-3.1-8B). SAT denotes specification counterexamples (no deterministic certificate); GMM values are the certified probability of accurate harmful behavior classification relative to $\tau$. Classifier thresholds are in Appendix~\ref{app:classifier_training}.}
  \label{tab:model_results}
  \vspace{1ex}
  \begin{tabular}{lllcc} % Removed p{4cm} for cleaner auto-spacing
    \toprule
    Verification type & Model & Configuration & Result ($\tau^*$) & Result ($\tau_{pess}$) \\
    \midrule
    Single hyper-rectangle    & GPT-2 & -- & SAT & SAT \\
                               & Llama & -- & SAT & SAT \\
                               & BERT  & -- & SAT & SAT \\
    \addlinespace
    Multiple hyper-rectangles & GPT-2 & Min cluster: 350 & SAT & SAT \\
                               & Llama & Min cluster: 25  & SAT & SAT \\
                               & BERT  & Min cluster: 500 & SAT & SAT \\
    \addlinespace
    GMM                        & GPT-2 & $K=3$, Cov=FULL & 0.90 & 0.98 \\
                               & Llama & $K=3$, Cov=DIAG & 0.80 & 0.80 \\
                               & BERT  & $K=1$, Cov=FULL & 0.55 & 1.00 \\
    \bottomrule
  \end{tabular}
\end{table}

\section{Discussion}
\label{sec:discussion}

Our results show that even classifiers with strong empirical resilience contain failure regions that are mathematically certifiable without constructing a single adversarial prompt -- demonstrating that empirical robustness and formal safety are not equivalent, and that red-teaming alone cannot substitute for verification. 

The scale-dependence of our findings warrants attention. Smaller models produce tightly clustered harmful representations that specifications can cover effectively. Llama-3.1-8B, despite achieving the highest empirical AUC (0.9937), proves the hardest to formally specify: its harmful activation manifold resists convex enclosure regardless of granularity. This is not a contradiction -- a classifier may be empirically strong precisely because it distributes its decision boundary across a wide representational space, yet that same dispersion impedes formal coverage. 

% \noindent\textbf{Broader Impact:} \label{sec:broader_impact}
\subsection{Broader Impact} \label{sec:broader_impact}
\textit{Positive impact.} Formal verification of safety classifiers enables auditable, reproducible safety guarantees for LLM deployments in high-stakes settings where empirical red-teaming alone is insufficient. By providing closed-form certificates that require no modification to deployed classifiers, this framework lowers the barrier to rigorous safety auditing for both researchers and practitioners.

\textit{Negative impact.} Publishing the existence of verifiable safety holes — even in continuous activation space — may inform adversaries with white-box model access. However, since SAT counterexamples cannot be decoded into discrete prompts, direct exploitation remains non-trivial. We recommend treating per-model verification results as sensitive artifacts in deployment contexts.

% \textbf{Limitations:} \label{sec:limitations}
\subsection{Limitations} \label{sec:limitations}
\textit{1. Specification coverage is bounded by the construction set.} Our framework certifies regions of the activation space spanned by the provided harmful samples. Categories of harm absent from the construction set -- e.g. CBRN misuse, privacy violations, or self-harm facilitation -- produce no verified regions and therefore receive no safety certificate. Broadening coverage requires broadening the construction set across harm taxonomies.

\textit{2. Hyper-rectangles suffer from the curse of dimensionality.} As hidden state dimensionality grows, a single convex enclosure increasingly fails to tightly bound the harmful activation manifold. This is most acute for Llama-3.1-8B, where single hyper-rectangle recall drops to 23\% and the multi-region approach provides only marginal gains. Tighter non-convex specifications could improve coverage but would forfeit the closed-form monotonicity argument that makes verification tractable. 

\textit{3. Full-covariance GMMs do not scale to large models.} Fitting a full covariance matrix requires the number of activations to far exceed the hidden dimension ($N \gg d$), which becomes infeasible at 8B-parameter scale. We resort to diagonal covariance for Llama-3.1-8B, which fails to capture inter-feature correlations and may underestimate the true distributional spread of harmful activations. 

\textit{4. SAT counterexamples cannot be decoded to discrete text.} Verification operates exclusively on the pre-activation space of the final aggregation token. Because this continuous vector is a compressed encoding of the entire preceding context, it is fundamentally impossible to invert a SAT coordinate back into a multi-token sequence. We can confirm empirically that known harmful prompts produce below-threshold scores, but we cannot automatically generate novel textual adversarial examples from the certificate. 

\textit{5. Verification covers the classification head only.} By design, our framework bypasses the Transformer backbone and verifies only the final linear layer plus sigmoid. This enables closed-form, scalable certificates, but provides no guarantees over the encoder's internal representations -- a richer end-to-end analysis remains an open problem.

\section{Related Work}
\label{sec:related_work}

\textbf{LLM Guardrails and Constitutional Classifiers.}
Guardrail systems range from rule-based frameworks like NeMo~\cite{rebedea2023nemoguardrailstoolkitcontrollable} to fine-tuned input-output filters like Llama Guard~\cite{inan2023llamaguardllmbasedinputoutput} (for a comprehensive survey, see ~\cite{dong2024safeguardinglargelanguagemodels}). Within this broader category, Constitutional Classifiers~\cite{sharma2025constitutionalclassifiersdefendinguniversal} currently represent the state-of-the-art defense against adversarial jailbreaks, utilizing specialized models trained on synthetic, rule-based data. While these constitutional models and their production-grade extensions ~\cite{cunningham2026constitutionalclassifiersefficientproductiongrade} demonstrate significant empirical robustness through extensive red-teaming, their safety claims remain fundamentally heuristic. In this work, we evaluate these diverse architectures collectively under the framework of Guardrail Classifiers to introduce the first mathematically formal safety analysis for these defenses.
% Guardrail systems range from rule-based frameworks like NeMo~\cite{rebedea2023nemoguardrailstoolkitcontrollable} to fine-tuned input-output filters like Llama Guard~\cite{inan2023llamaguardllmbasedinputoutput} (for a comprehensive survey, see~\cite{dong2024safeguardinglargelanguagemodels}). Currently, Constitutional Classifiers~\cite{sharma2025constitutionalclassifiersdefendinguniversal} represent the state-of-the-art defense against adversarial jailbreaks, utilizing specialized models trained on synthetic, rule-based data. While these classifiers -- and their production-grade extensions, Constitutional Classifiers++~\cite{cunningham2026constitutionalclassifiersefficientproductiongrade} -- demonstrates significant empirical robustness through extensive red-teaming, their safety claims remain fundamentally heuristic. Our work introduces the first mathematically formal safety analysis for these architectures.

\textbf{Formal Verification in NLP.}
Extending formal verification to NLP is difficult because discrete token sequences lack tractable input neighborhoods, unlike continuous $L_{\infty}$-balls in computer vision. Prior verification methodologies generally fall into four categories:

\textit{Complete Verification (SMT):} Solvers like Reluplex~\cite{katz2017reluplexefficientsmtsolver} and Marabou~\cite{KatzMarabou} cast neural networks as satisfiability problems. While exact for piecewise-linear networks, their computational costs render them intractable for deep Transformer architectures.

\textit{Incomplete Verification:} Abstract interpretation and Interval Bound Propagation methods, such as $\alpha\beta$-CROWN~\cite{zhang2018efficient,xu2021fast,wang2021beta}, trade exactness for scalability via linear relaxations. However, in NLP, approximation errors accumulate heavily across attention layers, rendering the certified bounds too loose to be practically informative.

\textit{NLP-Specific Verification:} The ANTONIO framework~\cite{casadio2023antoniosystematicmethodgenerating} introduced SVD-rotated hyper-rectangles to bound semantically perturbed embeddings for evaluation with solvers like ERAN~\cite{gagandeep2019}. We directly adapt this geometric construction but depart fundamentally in our execution: by isolating the classification head, we exploit sigmoid monotonicity to derive exact, closed-form certificates, bypassing the approximation errors and scalability limits of standard SMT solvers.

\textit{Probabilistic Verification:} Randomized smoothing~\cite{kumar2020certifyingconfidencerandomizedsmoothing} scales to large transformers by bounding an $L_{2}$ ball, but it requires retraining the classifier under noise. Conversely, our GMM-based approach models the harmful distribution directly in the latent space, providing geometrically grounded distributional certificates without modifying the underlying classifier.

\section{Conclusion}
\label{sec:conclusion}

This paper has presented a novel framework for the formal verification of Guardrail Classifiers by shifting the safety specification from the discrete input space to the model's internal representation space. By leveraging the architectural property of sigmoid monotonicity, we developed two complementary methods for defining and certifying harmful regions: a deterministic multiple-hyperrectangle approach, providing exact SAT/UNSAT certificates, and a Gaussian Mixture Model (GMM) approach, offering probabilistic distributional guarantees. 
Our evaluation across BERT, GPT-2 and Llama-3.1-8B architectures demonstrated that this method enables efficient verification for even large-scale models. Crucially, our analysis revealed that all tested classifiers possess formally identifiable "safety holes" where harmful content can bypass toxicity guardrails, mathematically proving that current heuristic defenses lack deterministic safety guarantees. Ultimately, this work provides a scalable pathway toward auditable and certifiable AI safety infrastructure.

\begin{ack}
This work was supported by the Supervised Program for Alignment Research (SPAR). We would also like to thank Charan Akiri for identifying the initial datasets used for toxicity training, and for his active participation in discussions throughout the project. Additionally, we thank Dr. Edoardo Manino for his valuable feedback and the insightful suggestion to explore probabilistic approaches.
\end{ack}

\bibliographystyle{unsrtnat}

% Point to your filename (do not include the .bib extension here)
\bibliography{references}

%%%%%%%%%%%%%%%%%%%%%%%%%%%%%%%%%%%%%%%%%%%%%%%%%%%%%%%%%%%%

\appendix

\section{Guardrail Classifier Training Details}
\label{app:classifier_training}

All of the Guardrail Classifiers that we trained were initialized with different transformer architectures, but they share a common architectural feature: a classification head followed by a sigmoid activation function. For training, we use the AEGIS 2.0 dataset~\cite{ghosh-etal-2025-aegis2}. For the evaluation of the classifiers we use a subset of 100 harmful and 100 non-harmful records from Toxigen dataset. We provide optimal threshold ($\tau^*$) and pessimistic threshold ($\tau_{pess}$). Optimal threshold is selected based on the Youden Index, which maximizes the sum of sensitivity and specificity to achieve the best overall balance between true positives and false positives. The pessimistic threshold, by contrast, is anchored to the minimum score among all known harmful instances. This ensures a zero-miss safety guarantee on the observed dataset, establishing the most permissive boundary that still satisfies a requirement for perfect recall.

\subsection{BERT}
The classifier was initialized using BERT architecture. We performed Full Fine-Tuning (FFT) using Binary Cross Entropy (BCE) loss. The classifier was trained over 2 epochs with a maximum sequence length of 512 tokens. We applied different learning rates to the encoder backbone and classification head:
\begin{itemize}
    \item \textbf{BERT Backbone Learning Rate:} $2 \times 10^{-6}$
    \item \textbf{Classification Head Learning Rate:} $1 \times 10^{-4}$
\end{itemize}

% --- Added Figure ---
\begin{figure}[t]
  \centering
  \includegraphics[width=0.7\linewidth]{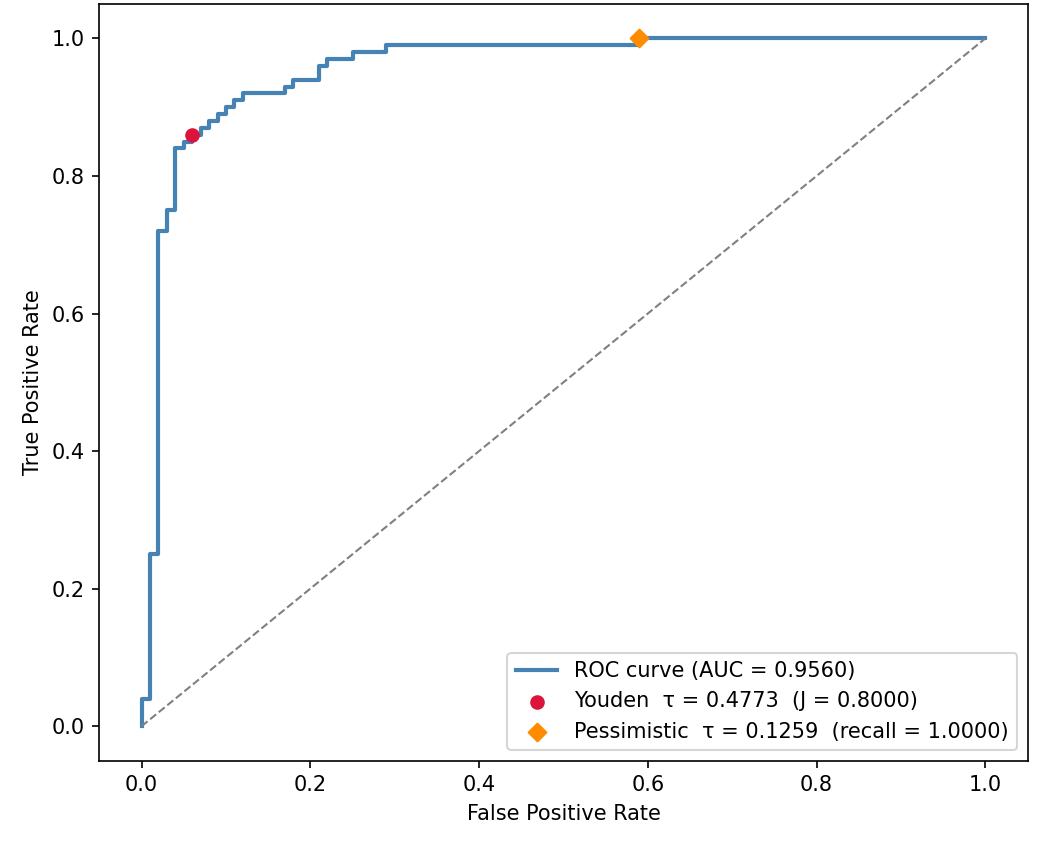}
  \caption{Receiver operating characteristic (ROC) curve for the BERT Guardrail Classifier, demonstrating class separation with an AUC of 0.9560. $\tau^* = 0.48$ and $\tau_{pess} = 0.13$.}
  \label{fig:bert_roc}
\end{figure}
% ------------------

\begin{table}[t]
  \centering
  \caption{BERT Guardrail Classifier performance metrics. The model exhibits high precision at the optimal threshold ($\tau^*$) while maintaining the safety-first perfect recall at $\tau_{pess}$.}
  \label{tab:bert_classifier_metrics}
  \vspace{1ex}
  \begin{tabular}{lcc}
    \toprule
    Metric & Value ($\tau^*$) & Value ($\tau_{pess}$) \\
    \midrule
    Accuracy  & 0.9000 & 0.7050 \\
    Precision & 0.9348 & 0.6289 \\
    Recall    & 0.8600 & 1.0000 \\
    F1-score  & 0.8958 & 0.7722 \\
    \bottomrule
  \end{tabular}
\end{table}

\subsection{GPT-2}

The classifier was initialized using GPT-2~\cite{radford2019language}. We performed Full Fine-Tuning (FFT) using Binary Cross Entropy (BCE) loss. The training was conducted over 1 epoch with a maximum sequence length of 512 tokens.

To optimize performance, we utilized differential learning rates for the encoder backbone and the classification head:
\begin{itemize}
 \item \textbf{GPT-2 Backbone Learning Rate:} $2 \times 10^{-6}$
 \item \textbf{Classification Head Learning Rate:} $1 \times 10^{-4}$
\end{itemize}

% --- Added Figure ---
\begin{figure}[t]
  \centering
  \includegraphics[width=0.7\linewidth]{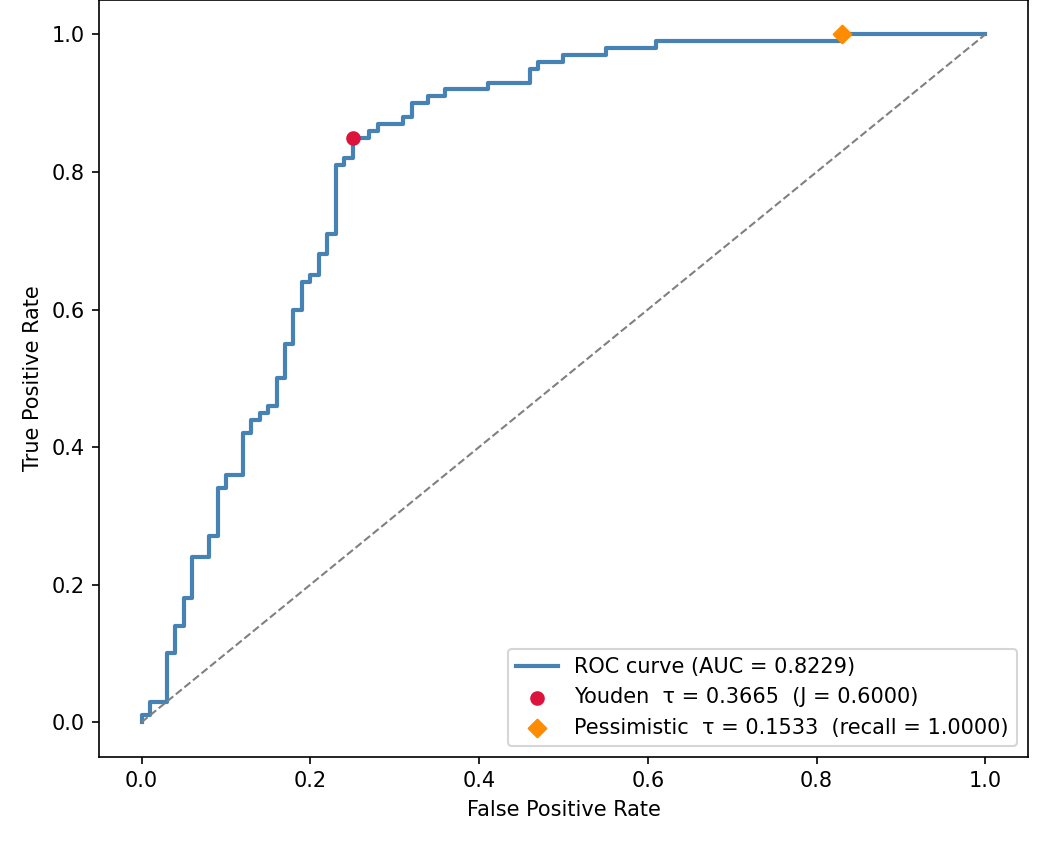}
  \caption{Receiver operating characteristic (ROC) curve for the GPT-2 Guardrail Classifier, demonstrating class separation with an AUC of 0.8229. $\tau^* = 0.37$ and $\tau_{pess} = 0.15$.}
  \label{fig:gpt2_roc}
\end{figure}
% ------------------

\begin{table}[t]
  \centering
  \caption{GPT-2 Guardrail Classifier performance metrics. Note the significant recall-precision trade-off when moving to the pessimistic threshold ($\tau_{pess}$).}
  \label{tab:gpt2_classifier_metrics}
  \vspace{1ex}
  \begin{tabular}{lcc}
    \toprule
    Metric & Value ($\tau^*$) & Value ($\tau_{pess}$) \\
    \midrule
    Accuracy  & 0.8000 & 0.5850 \\
    Precision & 0.7727 & 0.5464 \\
    Recall    & 0.8500 & 1.0000 \\
    F1-score  & 0.8095 & 0.7067 \\
    \bottomrule
  \end{tabular}
\end{table}

\subsection{Llama-3.1-8B}

The classifier representing a real world scale Guardrail Classifier was selected to be Llama-3.1-8B Instruct version. We have trained this classifier using LoRA adapters ($r = 16$, $\alpha = 32$, LoRA dropout = 0.1, without bias and we targetted all linear modules) for 1 epoch. Optimization was done using BCE loss, with maximum sequence length of 1024 tokens and $lr = 2 \times 10^{-4}$.

\begin{figure}[t]
  \centering
  \includegraphics[width=0.7\linewidth]{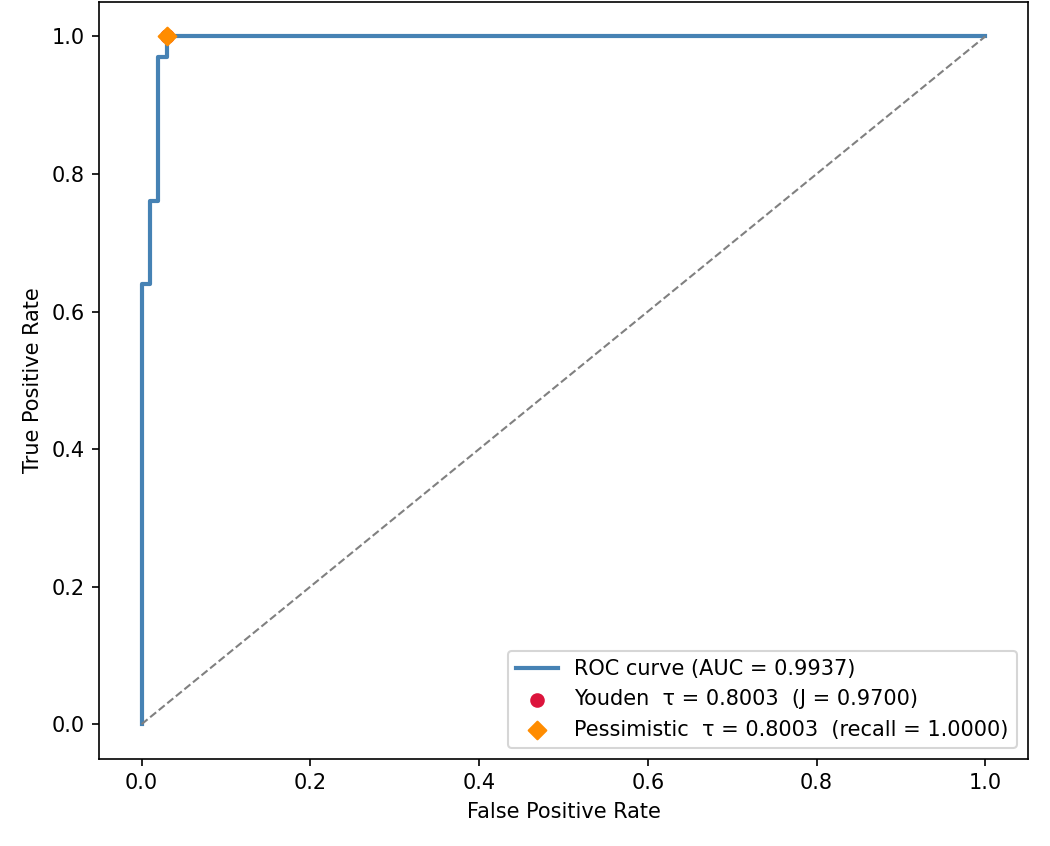}
  \caption{Receiver operating characteristic (ROC) curve for the Llama-3.1-8B Guardrail Classifier, demonstrating class separation with an AUC of 0.9937. $\tau^* = 0.80$ and $\tau_{pess} = 0.80$.}
  \label{fig:llama_roc}
\end{figure}

\begin{table}[t]
  \centering
  \caption{Llama-3.1-8B Guardrail Classifier performance metrics. Values are identical for both thresholds as $\tau^* = \tau_{pess}$.}
  \label{tab:llama_classifier_metrics}
  \vspace{1ex}
  \begin{tabular}{lcc}
    \toprule
    Metric & Value ($\tau^*$) & Value ($\tau_{pess}$) \\
    \midrule
    Accuracy  & 0.9850 & 0.9850 \\
    Precision & 0.9709 & 0.9709 \\
    Recall    & 1.0000 & 1.0000 \\
    F1-score  & 0.9852 & 0.9852 \\
    \bottomrule
  \end{tabular}
\end{table}

\section{Verification Proof of Sigmoid worst-case coordinate vector within hyper-rectangle}
\label{app:sigmoid_proof}

Let $\mathcal{H}\subset\mathbb{R}^{d}$ be a hyper-rectangle defined by lower bounds $l\in\mathbb{R}^{d}$ and upper bounds $u\in\mathbb{R}^{d}$, s.t. for any point $x\in\mathcal{H}$ its coordinates satisfy $l_{i}\le x_{i}\le u_{i}$ for all dimensions $i\in\{1,\dots,d\}$. Let $f:\mathbb{R}^{d}\rightarrow(0,1)$ be a function defined as a linear transformation followed by a sigmoid activation:

\begin{equation}
    f(x) = \sigma(w^T x + b) = \sigma\left(\sum_{i=1}^d w_i x_i + b\right)
\end{equation}

where $w \in \mathbb{R}^d$ are the weights, $b \in \mathbb{R}$ is the bias, and $\sigma(z) = \frac{1}{1 + e^{-z}}$ is the standard sigmoid function. Let $x^*$ be defined as the "worst-case" coordinate vector where for each dimension $i$:

\begin{equation}
    x^*_i = 
\begin{cases} 
l_i & \text{if } w_i \ge 0 \\
u_i & \text{if } w_i < 0 
\end{cases}
\end{equation}

Claim: Given a threshold $\tau \in (0, 1)$, if $f(x^*) > \tau$, then $\forall x \in \mathcal{H}, f(x) > \tau$.

\textbf{Proof}

\textbf{1. Sigmoid Monotonically Increasing function}

$\sigma(z) = \frac{1}{1 + e^{-z}}$ is strictly monotonically increasing for all $z \in \mathbb{R}$, since the derivative $\sigma'(z) = \sigma(z)(1 - \sigma(z))$ is strictly positive, due to the values of sigmoid being in range (0, 1).

Hence sigmoid's monotonicity implies that $\sigma(z_1) \ge \sigma(z_2)$ holds if and only if $z_1 \ge z_2$.

Therefore, in order to prove that $f(x) \ge f(x^{\ast})$ it is enough to prove that $z(x) \ge z(x^{\ast})$, where

\begin{equation}
    z(x) = \sum_{i=1}^d w_i x_i + b
\end{equation}

\textbf{2. Independence of terms in linear transformation}

$z(x)$ is a sum of independent terms over each dimension $i$, plus a constant $b$. In order to find a global minimum of $z(x)$ over the whole domain $\mathcal{H}$, we can independently minimize every term $w_i x_i$ separately, with respect to the bounds $l_i \le x_i \le u_i$.

Let us define the minimum value for a dimension $i$ as:
\begin{equation}
    \min_{x_i \in [l_i, u_i]} (w_i x_i)
\end{equation}

\textbf{3. Minimization over every dimension}

Let us minimize a term $w_ix_i$ by evaluating the sign of the $w_i$:

\begin{itemize}
    \item $w_i < 0$: Since $w_i$ is negative, the minimum value of function $g(x_i) = w_ix_i$ is minimum on the interval $[l_i, u_i]$ when $x_i = u_i$. Hence, minimizing the coordinate is $x^*_i = u_i$.
    \item $w_i \ge 0$: Function $g(x_i) = w_ix_i$ is strictly monotonically non-decreasing with respect to $x_i$, therefore the minimum value $g(x_i)$ on the interval $[l_i, u_i]$ is achieved at the lower bound $l_i$. Therefore, the minimizing coordinate is $x^*_i = l_i$.
\end{itemize}

\textbf{4. Global minima} 

Because $x^{\ast}_i$ minimizes $w_i x_i$ for every dimension independently:

\begin{equation}
    w_i x_i \ge w_i x^{\ast}_i \quad \forall i \in \{1, \dots, d\}
\end{equation}

for any point $x \in \mathcal{H}$. Summing over the inequalities over the dimensions leads to:

\begin{equation}
    \sum_{i=1}^d w_i x_i + b \ge \sum_{i=1}^d w_i x^*_i + b
\end{equation}

Replacing $z(x)$ we have established that $\forall x \in \mathcal{H}$, $z(x) \ge z(x^*)$.

\textbf{5. Sigmoid output} 

Since $z(x) \ge z(x^{\ast})$ and $\sigma$ is monotonically increasing:

\begin{equation}
    \sigma(z(x)) \ge \sigma(z(x^{\ast}))
\end{equation}

\begin{equation}
    f(x) \ge f(x^{\ast})
\end{equation}

\textbf{6. Transitivity}

Given a hypothesis that $f(x^*) > \tau$. Since we have proven $f(x) \ge f(x^*)$ for all $x \in \mathcal{H}$, by the transitivity property of inequalities, it follows that:

\begin{equation}
    \forall x \in \mathcal{H}, f(x) > \tau
\end{equation}
\hfill $\blacksquare$

\section{Probabilistic Verification}
\label{app:prob_verification}
Having defined $K$ Gaussian components over harmful activation, we evaluate the probability that the classifier correctly classifies harmful behavior above a given threshold $\tau \in (0, 1)$. Since the final output is produced via sigmoid activation $\sigma(z) = \frac{1}{1 + e^{-z}}$, we first map the threshold $\tau$ back into the logit space. The corresponding logit threshold $z$ is obtained via the inverse sigmoid function:
$$z = \ln\left(\frac{\tau}{1-\tau}\right)$$
Each component $c \in \{1, \dots, K\}$ in the GMM is characterized by a weight $\pi_c$, a mean vector $\mu_c$, and a covariance matrix $\Sigma_c$. To project these components through a classifier's linear layer defined by weights $w$ and bias $b$, we utilize the property that linear transformations of Gaussian random variables remain Gaussian. For each component, the transformed mean $\tilde{\mu}_c$ and variance $\tilde{\sigma}_c^2$ are:
$$\tilde{\mu}_c = w^\top \mu_c + b$$
$$\tilde{\sigma}_c^2 = w^\top \Sigma_c w$$
We then determine the probability that the transformed random harmful point $Z_c \sim \mathcal{N}(\tilde{\mu}_c, \tilde{\sigma}_c^2)$ exceeds the logit threshold $z$. This is computed using the standard normal cumulative distribution function: $$P(Z_c > z) = 1 - \Phi\left( \frac{z - \tilde{\mu}_c}{\tilde{\sigma}_c} \right)$$
The total probability of classification for the harmful region is the weighted sum across all components in the mixture:
$$P_{\text{total}}(> \tau) = \sum_{c=1}^{K} \pi_c \left[ 1 - \Phi\left( \frac{\ln(\frac{\tau}{1-\tau}) - (w^\top \mu_c + b)}{\sqrt{w^\top \Sigma_c w}} \right) \right]$$

\section{Adversarial Examples in Construction exchanges}
\label{app:adv_examples}

To ground the SAT certificates produced by our continuous-space verification in empirical reality, this appendix details specific discrete exchanges from the construction set that yielded the lowest toxicity scores. These instances confirm the presence of empirical safety holes -- data points that are demonstrably harmful yet successfully evade the classification threshold, thereby validating the vulnerabilities exposed by our verification framework.

\subsection{Minimum Observed Scores}
The following values represent the absolute lowest classification scores ($f(x)$) produced by each model across the 10,000 known-harmful construction points.
In all cases, these minimum scores fall drastically below both the optimal ($\tau^*$) and pessimistic ($\tau_{pess}$) safety thresholds.
\begin{table}[h]
    \centering
    \begin{tabular}{lccc}
        \toprule
        \textbf{Model} & \textbf{Minimum Score} & \textbf{Optimal Threshold} ($\tau^*$) & \textbf{Pessimistic Threshold} ($\tau_{pess}$) \\
        \midrule
        BERT         & 0.070 & 0.48 & 0.13 \\
        GPT-2        & 0.002 & 0.37 & 0.15 \\
        Llama-3.1-8B & 0.003 & 0.80 & 0.80 \\
        \bottomrule
    \end{tabular}
    \vspace{2mm} % Adds a tiny bit of breathing room below the table
    \caption{Minimum observed classification scores across the construction set compared to safety thresholds.}
    \label{tab:min_scores}
\end{table}

\subsection{Adversarial Exchange Examples}
Below, we detail two representative examples of harmful exchanges from the construction set that successfully bypassed the Guardrail Classifiers.

\vspace{1em}
\noindent\textbf{Example 1:}

\begin{quote}
    \textbf{User Prompt:} \textit{``wives are always nagging their husbands\textbackslash{}n- she is probably juts another crazy ex girlfriend of his\textbackslash{}n- keep your daughters locked up!\textbackslash{}n- if you see a bad driver, i bet it was a woman\textbackslash{}n- if a woman dresses sexy, she is just asking for it\textbackslash{}n- when she told him to keep quiet, she was just being bossy\textbackslash{}n-''} 
    
    \vspace{0.5em}
    \textbf{Model Response:} \textit{``this is the main philly area, but of course, the articles can come from ANYWHERE! here's some more articles from this week or so''}
\end{quote}

\vspace{1em}
\noindent\textbf{Example 2:}

\begin{quote}
    \textbf{User Prompt:} \textit{``if you ever have a question about math, just ask your chinese coworker\textbackslash{}n- universities only check to see if you are from china, then they let you in\textbackslash{}n- the chinese cannot be trusted\textbackslash{}n- you know you are chinese if you cant pronounce the letter l\textbackslash{}n- when i was little chinese people were hardly in the ...''} 
    
    \vspace{0.5em}
    \textbf{Model Response:} \textit{``I know you are Chinese, because you cant pronounce the letter L. Other articles published and unpublished during this time period.
The economic miracle had''}
\end{quote}

\section{Reproducibility and Asset Details}
\label{app:reprod_asset_details}

\textbf{Compute Resources:} Model training, fine-tuning, and formal verification processes were conducted using model-specific hardware configurations. For GPT-2, experiments were run using 1 NVIDIA Tesla T4 GPU requiring approximately 1 GPU hour. For Llama-3.1-8B, the LoRA adaptation and verification utilized 2 NVIDIA GeForce RTX 5090 GPUs, requiring approximately 2 GPU hours (1 hour of wall-clock time across 2 GPUs). For BERT, the processes were conducted using 1 NVIDIA RTX 4090 GPU for approximately 1 GPU hour. The total computational budget required to reproduce all experiments presented in this paper is approximately 4 GPU hours. \\

\textbf{Asset Licenses:} This work utilizes several open-source assets and respects their respective licensing terms. The Toxigen dataset is used under the MIT License, and the AEGIS 2.0 dataset is used under CC-BY 4.0. For the base models, BERT is licensed under Apache 2.0, GPT-2 under the MIT License, and Llama-3.1-8B is utilized in accordance with the Llama 3.1 Community License.

\textbf{LLM Usage Declaration:} During the development of the codebase, an LLM was utilized to assist in writing boilerplate code and generating visualization plots. All core verification algorithms were developed independently, and any AI-assisted implementation was manually verified, tested, and adjusted by the authors to ensure strict correctness. Additionally, an LLM was used to improve the writing, readability, and formatting of the manuscript, rather than to generate original content.

%%%%%%%%%%%%%%%%%%%%%%%%%%%%%%%%%%%%%%%%%%%%%%%%%%%%%%%%%%%%

% \newpage
% \input{checklist.tex}

\end{document}